\documentclass[conference]{IEEEtran}
\IEEEoverridecommandlockouts

\usepackage{amsmath,amssymb}
\usepackage{booktabs}
\usepackage{array}
\usepackage{multirow}
\usepackage{graphicx}
\usepackage{url}
\usepackage{cite}
\usepackage{balance}
\usepackage{tikz}
\usetikzlibrary{arrows.meta,positioning,fit,shapes.geometric,patterns}
\usepackage{pgfplots}
\pgfplotsset{compat=1.18}
\usepackage{xcolor}
\usepackage[hidelinks]{hyperref}

\definecolor{grayA}{gray}{0.18}
\definecolor{grayB}{gray}{0.55}
\definecolor{grayC}{gray}{0.88}

\title{SerenAI: State-transition system inspired by text-based world AI models}

\author{\IEEEauthorblockN{Elvin Babayev, Artem Sinitsa, Arash Hajisharifi, Kabir Bakhshaei}
\IEEEauthorblockA{Collision Technologies S.R.L.S. Societa Benefit\\
Trieste, Italy\\
company@collitech.org}}

\begin{document}
\maketitle

\noindent\textit{Abstract:}
Although professional workflows leverage large language models widely, the interpretation for auditing unconstrained free-text generation is usually intractable if such generation demands legal, operational or financial workflow. We hereby demonstrate a text based system called SerenAI - inspired by world-models, it is a state transition system that outputs verifiable predictions rather than merely text: Provided with a description of the environment, state, and actions, the generated output contains 4 items: causal deltas that causally effect the given state, a next state that can logically follow from the given state and action, a validity reward, and a termination signal. For the released proto-model, we employ 2 steps of adaptation training, namely parameter efficient fine-tuning followed by verifier based RL over 50,000 exampled cause and effects in 12 environments spanning 10 reasoning domains. Compared to an initial internal evaluation of an 8B open-weight baseline, SerenAI increased JSON validity from 85.0\% to 93.2\%, schema validity from 55.0\% to 84.0\%, exact structured-output match from 0.0\% to 41.5\%, causal-delta exact match from 0.0\% to 41.5\%, resulting-state exact match from 0.0\% to 42.0\%, reward exact match from 1.0\% to 80.5\%, and termination exact match from 38.0\% to 81.5\%. These support the narrower claim that verifier-compatible adaptation can improve structured transition prediction.  They do not yet establish legal-grade reliability. Accordingly, the paper also specifies a validation protocol for evidence-grounded legal workflows, calibration, human oversight, and sovereign on-premise deployment.


\section{Introduction}
Generative language models can write, summarize and help retrieval – but the just predictive objective function is unable to guarantee they are factual, internally consistent, or compliant. When hallucinations enter into higher stakes operations, missing due dates, non-supported regulatory claim, mis-referenced legislation, or inserting errant clauses are often significant professional failures. The increasing literature already shows numerous types of hallucination and many grounding, constrained generation, verification and proportional human check techniques as the way to tackle them, not a few elicitation prompts \cite{huang2023hallucination,lewis2020rag}.

The Collision Technologies research  covers the risk of hallucination, safe deployment and AI use cases. To achieve this goal in practice, SerenAI targets a “structured-form” reasoning approach that’s well-suited for private and regulated environments. We aim to focus our initial commercial product efforts on legal work – such as performing reviews of documents or compliance, monitoring deadlines for processes and judicial documents or providing specialized advice on legal discovery. For now, the technical documentation includes the cause and effect reasoning is based on synthetic or collected environments but doesn’t yet validate it for legal advice.

The system is inspired by world-model research, which studies predictive internal models of environment dynamics \cite{ha2018worldmodels,lecun2022path}. Classical world models often learn latent dynamics from sensor or interaction trajectories. SerenAI addresses a narrower textual setting. It treats a workflow as an action-conditioned state-transition problem and predicts an explicit symbolic transition. We therefore use the term \emph{world-model-inspired system} rather than claiming a general physical or multimodal world model.

The principal contributions of this paper are:
\begin{enumerate}
    \item A four-field transition representation in which we specify what changed, the next state, validity and completion;
    \item A two-stage training scheme in which parameter-efficient supervised fine-tuning is combined with an automatic structured-output verifier and reinforcement learning;
    \item a transparent report of preliminary internal benchmark results, including limitations that prevent a claim of legal-grade reliability; and
    \item A deployment \& validation framework for sovereign, evidence based AI solutions in regulated professional practice.
\end{enumerate}

\section{Related Work}
\subsection{World models describe the dynamics of an environment in a way that enables the agent to predict, plan, or control the environments.} 
Ha and Schmidhuber compressed the representations for spatial and temporal structures of environments to reinforcement-learning environments. \cite{ha2018worldmodels}. LeCun later proposed predictive architectures in which agents learn abstract representations of the world and reason through internal models rather than relying exclusively on reactive pattern matching \cite{lecun2022path}. SerenAI adopts the state-transition principle but represents the state and action in text and requires the next-state prediction to be emitted as a verifiable schema.

SerenAI uses state transition rule and states and actions are presented in text and the next-state prediction is emitted as a verifiable schema.
This distinction is important. A language model that produces a smooth text might not maintain entity identity, temporal ordering, preconditions or termination conditions.

In contrast, in the transition model the evaluation is against the specific variables. However, it is closer to symbolic workflow simulation than generic dialogue, but as in all else, a pretrained language model is still used as the function approximator.

\subsection{Efficient Adaptation and Verifier-Guided Learning}
High-quality instruction data can substantially boost response LIMA has demonstrated that a small number of carefully chosen examples can equip models with robust instruction-following and interaction behavior \cite{zhou2023lima}.Parameter-efficient approaches like Quantized Low-Rank Adaptation (QLoRA) achieve memory efficient adaptation by training low-rank adapters on a frozen quantized backbone \cite{dettmers2023qlora}. They can be  excellent choices  for sovereign AI, since they keep costs of training and deployment low while still maintaining control of the weights and data.

Verification gives a second source of supervision. Cobbe et al. also demonstrated trained verifiers can enhance mathematical reasoning when ranking or choosing model completions \cite{cobbe2021verifiers}. DeepSeekMath proposes Group Relative Policy Optimization (GRPO) for sample-efficient reinforcement learning of sequential reasoning behaviors \cite{shao2024deepseekmath}. SerenAI employs the same overarching pattern: the structure of our task outputs allows for automated verification, which in turn provides an easily scalable reward signal. The paper makes no claim to novel policy-optimization algorithms; its contribution is to fuse verifier-compatible state-transition outputs into a safe-guarded workflow system.

\subsection{Structured generation, grounding, and legal AI}

Structured generation can be supported in many cases with training under a schema-aware objective, or what is know as “constrained decoding”. A frequent method, and utilized in PICARD, is to assign large negative scores to any invalid token during the process of auto regressive generation for text-to-SQL. \cite{scholak2021picard}; other work uses grammar-constrained decoding to enforce syntax with respect to a formal output language entirely \cite{geng2023grammar}. The SerenAI approach only reports learned adherence after fine-tuning and reinforcement learning. In production, learned adherence could simply be combined with constrained decoding as a two-pronged approach.

RAG applies parametric models jointly with explicit non-parametric evidence for knowledge intensive tasks \cite{lewis2020rag}. In legal AI, LegalBench offers 162 tasks testing several kinds of legal reasoning \cite{guha2023legalbench} and CUAD provides expert annotations for contract-review clauses \cite{hendrycks2021cuad}. Those benchmarks show that legal reliability requires task-level evaluation and expert consultation. A generic reasoning benchmark is not enough to ensure safe legal performance.

\section{Problem Formulation}
Let $e$ denote an environment, $s_t$ the current state at step $t$, $a_t$ an action or event, and $c_t$ optional context. SerenAI models the conditional transition
\begin{equation}
    p_{\theta}(y_t \mid e,s_t,a_t,c_t),
\end{equation}
where the output is a tuple
\begin{equation}
    y_t = (\Delta_t, s_{t+1}, r_t, d_t).
\end{equation}
Here, $\Delta_t$ is the causal delta describing what changed, $s_{t+1}$ is the resulting state, $r_t$ is a reward or validity indicator, and $d_t$ is a termination signal. The representation forces the model to separate transition content from task status.

A canonical output has the following form:
\begin{verbatim}
{
  "causal_delta": "...",
  "resulting_state": "...",
  "reward": 1,
  "done": false
}
\end{verbatim}
The schema enables deterministic validation of syntax, required keys, data types, and exact field values. For a target $y_t^*$, a verifier-compatible reward may be written as
\begin{equation}
R(y_t,y_t^*) = \sum_{k \in K} w_k \mathbb{1}[y_{t,k}=y_{t,k}^*]
- \lambda \mathbb{1}[y_t \notin \mathcal{Y}_{\mathrm{valid}}],
\end{equation}
where $K$ indexes the four fields and $\mathcal{Y}_{\mathrm{valid}}$ is the valid schema.

\section{System Architecture}
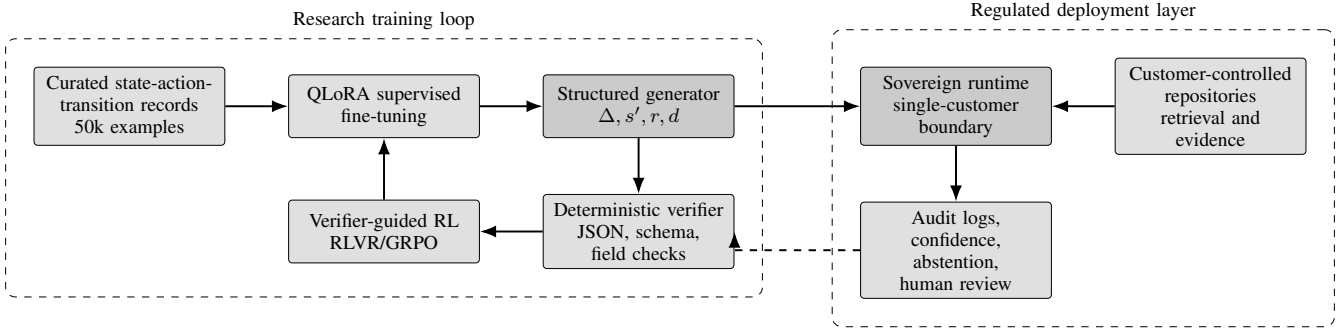
\begin{figure*}[t]
\centering
\resizebox{0.97\textwidth}{!}{%
\begin{tikzpicture}[
  node distance=8mm and 9mm,
  box/.style={draw=black, rounded corners=1.5pt, align=center, minimum height=9mm, text width=25mm, fill=grayC, font=\footnotesize},
  darkbox/.style={draw=black, rounded corners=1.5pt, align=center, minimum height=9mm, text width=25mm, fill=grayB!45, font=\footnotesize},
  line/.style={-Latex, thick},
  dashedline/.style={-Latex, thick, dashed}
]
\node[box] (data) {Curated state-action-transition records\\50k examples};
\node[box, right=of data] (sft) {QLoRA supervised fine-tuning};
\node[darkbox, right=of sft] (model) {Structured generator\\$\Delta, s', r, d$};
\node[box, below=of model] (verifier) {Deterministic verifier\\JSON, schema, field checks};
\node[box, left=of verifier] (rl) {Verifier-guided RL\\RLVR/GRPO};
\node[darkbox, right=18mm of model] (runtime) {Sovereign runtime\\single-customer boundary};
\node[box, right=of runtime] (retrieval) {Customer-controlled repositories\\retrieval and evidence};
\node[box, below=of runtime] (controls) {Audit logs, confidence, abstention, human review};

\draw[line] (data) -- (sft);
\draw[line] (sft) -- (model);
\draw[line] (model) -- (verifier);
\draw[line] (verifier) -- (rl);
\draw[line] (rl) -- (sft);
\draw[line] (model) -- (runtime);
\draw[line] (retrieval) -- (runtime);
\draw[line] (runtime) -- (controls);
\draw[dashedline] (controls.west) -| (verifier.east);

\node[draw=black, dashed, rounded corners, fit=(data)(sft)(model)(verifier)(rl), inner sep=4mm, label={[font=\footnotesize]above:Research training loop}] {};
\node[draw=black, dashed, rounded corners, fit=(runtime)(retrieval)(controls), inner sep=4mm, label={[font=\footnotesize]above:Regulated deployment layer}] {};
\end{tikzpicture}%
}
\caption{SerenAI research and deployment architecture. Solid arrows denote data or inference flow; the dashed return path denotes audit feedback from deployment into evaluation rather than automatic customer-data training.}
\label{fig:architecture}
\end{figure*}

Figure ~\ref{fig:architecture} outlines both a research and deployment architecture. The research phase involves a training setup that uses the curated state action-transaction data, which is going to be used for a supervised fine-tuning step that trains an output contract and the transition model. A deterministic verifier then is used to assesses structured output results and finally reinforcement learning utilize the feedback to increase the likelihood of outputs satisfying format requirements and target state conditions. On the deployment side, the model is isolated from customer-managed evidence stores, and the citations, confidence signals, logging and human-review gates are added.

The architecture built around the concept of data sovereignty. This allows sensitive prompts, files, intermediate stages, or responses to remain at all times on the premises without leaving the server. Local retrieval allows model responses to be anchored in trusted and customer-controlled repositories. Audit logs maintain records and model version for every response  produced. Use of customer data to train models should occur only when authority provides specific authorization, and separate consent.

\section{Dataset and Training Procedure}
\subsection{Curated transition corpus}
The study uses 50,000 cause-and-effect dataset distributed across 12 environments and 10 reasoning domains. This dataset includes formal logic, causal counterfactuals, chemistry, physics, number theory, spatial reasoning, and forward-chaining workflows as representative domains and each example contains an initial context, an action or event, and a target four-field transition. The data then was partitioned into  45,000 training, 2,500 validation, and 2,500 test records to execute supervised fine tuning.

\begin{table}[t]
\caption{Reported Corpus and Output Design}
\label{tab:corpus}
\centering
\footnotesize
\begin{tabular}{@{}p{0.31\columnwidth}p{0.62\columnwidth}@{}}
\toprule
\textbf{Item} & \textbf{Reported value} \\
\midrule
Total records & 50,000 \\
Training split & 45,000 \\
Validation split & 2,500 \\
Test split & 2,500 in split description; separate narrative says 5,000 evaluation samples \\
Environments & 12 \\
Reasoning domains & 10; seven examples named in this study\\
Output fields & causal delta, resulting state, reward, done \\
Training hardware & NVIDIA H100 80GB \\
Adaptation & QLoRA SFT followed by verifier-guided RL/GRPO experimentation \\
\bottomrule
\end{tabular}
\end{table}

\subsection{Two-Stage Adaptation}
The first phase conducts parameter efficient supervised fine-tuning method. As mentioned earlier quantized adaption is more applicable here because the backbone remains frozen in quantized form while low-rank adapters learn task-specific behavior \cite{dettmers2023qlora}. The aim of the supervised training objective is to minimize at the token level the negative log-likelihood of the structured target sequence. It is represented as follows:
\begin{equation}
    \mathcal{L}_{\mathrm{SFT}}(\theta) = -\sum_{i=1}^{N}\sum_{j=1}^{|y_i|}
    \log p_{\theta}(y_{i,j}\mid x_i,y_{i,<j}).
\end{equation}

\begin{itemize}
\item[a)]\textbf{$\mathcal{L}_{\text{SFT}}(\theta)$ (Total Error):} This indicates the amount of total errors the model makes with its current parameters ($\theta$).
    
\item[b)]\textbf{Minus sign ($-$) and logarithm ($\log$):} Since the probabilities given by the model are between 0 and 1, their logarithm gives a negative number. To measure the error as a positive "penalty score", a negative sign is placed in front of the formula.
    
\item[c)]\textbf{First summation $\sum_{i=1}^{N}$ (All Examples):} This means that the calculation is performed on all $N$ number of question-answer pairs in the training dataset and is summed.
    
\item[d)]\textbf{Second sum $\sum_{j=1}^{|y_i|}$ (All Words/Tokens):} This indicates that the words (tokens) in each answer of the model are traversed one by one. $|y_i|$ indicates how many tokens the answer consists of.
    
\item[e)] \textbf{$p_\theta(y_{i,j} \mid x_i, y_{i,<j})$ (Probability of Guessing the Next Word):} The most basic part of the formula is:
\begin{itemize}
        \item $x_i$: Question or task (prompt) given to the model.
        \item $y_{i,<j}$: The previous part of the correct answer given to the model up to that moment.
        \item $y_{i,j}$: The actual next correct word/token to be written.
\end{itemize}
\textbf{Explanation:} "Given the question and the correct words so far, what is the percentage probability that the model will choose the \textbf{exactly the next correct word}?"
\end{itemize}

\subsubsection*{Summary}

\begin{quote}
"We give the model a question and a portion of the correct answer. Then we check how likely the model is to guess the next correct word. If the model gives a low probability to the correct word, the formula gives it a high error (penalty) score. These probabilities are calculated for each word of each sentence in the entire database, summed up, and the model's overall error is found."
\end{quote}
Fig.2 shows plot of training loss dips rapidly at the beginning of the training and holds steady, as we would expect from fit to the target format that is well-structured. Rapid decline of the loss function indicates that how well model learns from curated data.
\begin{figure}[t]
\centering
\includegraphics[width=\columnwidth]{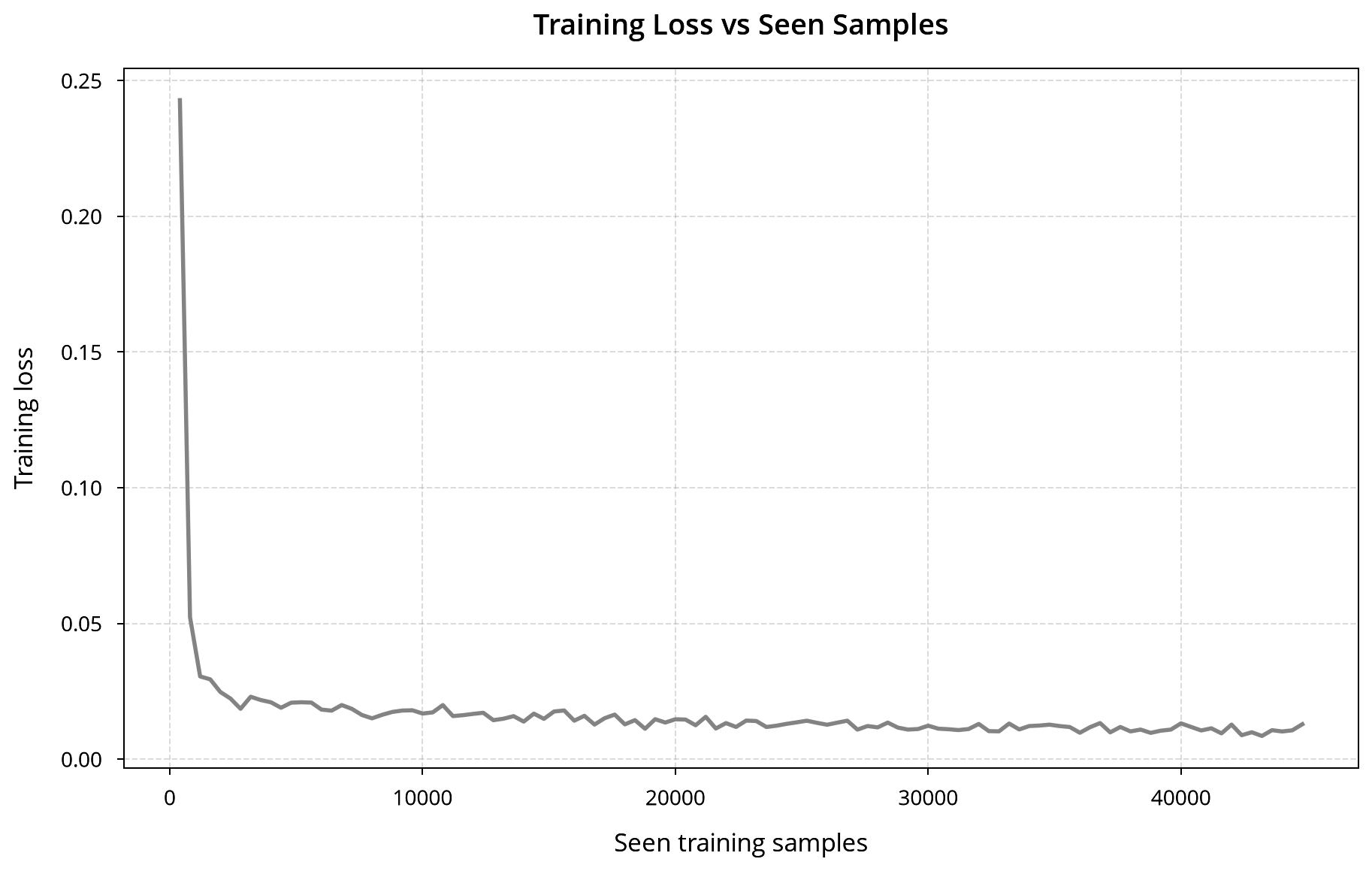}
\caption{Supervised-training loss function}
\label{fig:loss}
\end{figure}
The second stage uses the automatic verifier to score sampled outputs. The optimizer for the GRPO phase is AdamW. 
We set the learning rate as 5 × $10^{-6}$, weight decay to be 0.01  and perform gradient clipping with maximum norm at 1.0. For each prompt, eight candidate responses are sampled to form the GRPO  comparison group. Probability-ratio clipping is applied with $\epsilon$ = 0.2.

\section{Evaluation Methodology}
\subsection{Metrics}
The internal evaluation uses seven exact or validity-based metrics:
\begin{itemize}
    \item \textbf{JSON validity}: whether the output parses as JSON;
    \item \textbf{schema validity}: whether all four required fields and data types are present;
    \item \textbf{exact JSON match}: exact equality of the complete normalized output;
    \item \textbf{causal-delta exact}: exact match of the transition delta;
    \item \textbf{state exact}: exact match of the resulting state;
    \item \textbf{reward exact}: exact match of the validity/reward field; and
    \item \textbf{done exact}: exact match of the termination signal.
\end{itemize}
Exact match is explicitly strict by design, it can be automatically verified. It can sometimes undervalue text states which are semantically identical, further evaluation of semantic accuracy should include canonicalization, entity-level F1, graph or slot accuracy and expert-provided semantic evaluation.
\subsection{Preliminary Results}
\begin{figure*}[t]
\centering
\begin{tikzpicture}
\begin{axis}[
    width=0.96\textwidth,
    height=6.1cm,
    ybar,
    bar width=9pt,
    ymin=0,
    ymax=110,
    ylabel={Rate (\%)},
    symbolic x coords={JSON,Schema,ExactJSON,Delta,State,Reward,Done},
    xtick=data,
    xticklabels={JSON validity,Schema validity,Exact JSON,Causal delta,Resulting state,Reward,Done flag accuracy},
    x tick label style={rotate=22,anchor=east,font=\footnotesize},
    ymajorgrids=true,
    grid style={dashed,gray!40},
    legend style={at={(0.5,1.02)},anchor=south,legend columns=2,draw=none,font=\footnotesize},
    nodes near coords,
    nodes near coords style={font=\scriptsize},
    every axis plot/.append style={draw=black},
    enlarge x limits=0.07
]
\addplot+[fill=white,postaction={pattern=north east lines}] coordinates {
(JSON,85.0) (Schema,55.0) (ExactJSON,0.0) (Delta,0.0) (State,0.0) (Reward,1.0) (Done,38.0)};
\addplot+[fill=gray!45] coordinates {
(JSON,93.2) (Schema,84.0) (ExactJSON,41.5) (Delta,41.5) (State,42.0) (Reward,80.5) (Done,81.5)};
\legend{8B base model,SerenAI}
\end{axis}
\end{tikzpicture}
\caption{Reported preliminary benchmark results of SerenAI. Percentages are reproduced from the internal study of Collision Technologies team.}
\label{fig:benchmark}
\end{figure*}
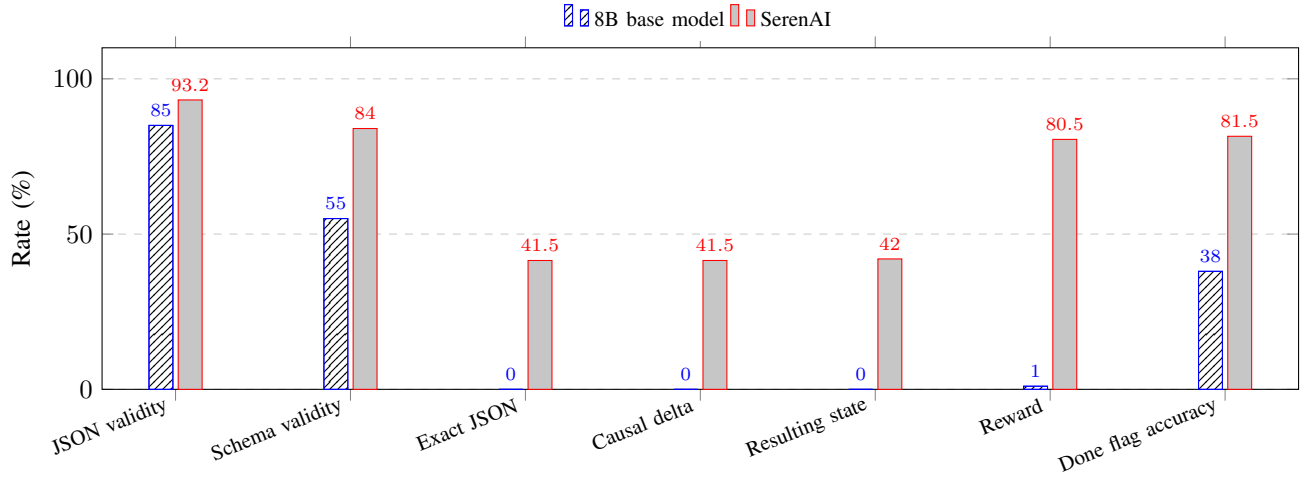

Fig.~\ref{fig:benchmark} and Table~\ref{tab:results} displays the percentages reported by this study

\begin{table}[t]
\caption{Reported Preliminary Results}
\label{tab:results}
\centering
\footnotesize
\begin{tabular}{lrrr}
\toprule
\textbf{Metric} & \textbf{Base} & \textbf{SerenAI} & \textbf{$\Delta$ pp} \\
\midrule
JSON validity & 85.0 & 93.2 & +8.2 \\
Schema validity & 55.0 & 84.0 & +29.0 \\
Exact JSON match & 0.0 & 41.5 & +41.5 \\
Causal delta exact & 0.0 & 41.5 & +41.5 \\
Resulting state exact & 0.0 & 42.0 & +42.0 \\
Reward exact & 1.0 & 80.5 & +79.5 \\
Done exact & 38.0 & 81.5 & +43.5 \\
\bottomrule
\end{tabular}
\end{table}

The validity metrics indicates a better machine-readability of the output. The main impact, however, comes from the increase of field-level exact matches. 
A model which only learned the format of the prediction might improve JSON and schema validity without increasing values of casual delta/resulting state.
Our observed improvements are to roughly 42\% and both transition fields are consistent with this, therefore this can clearly be described as task-learning evidence on an internal benchmark. However, our internal benchmark cannot distinguish memorization, rules with limited applicability, domain transfer learning, and robustness to adversarial attacks.

The largest absolute improvement comes from the structured fields: the reward and termination fields. These two fields perhaps being only marginally more complicated than the state prediction on a free text field can also be affected by class imbalances. Looking at the class distributions and using majority baselines, 80.5\% and 81.5\% are seen to be effective exact-field accuracies.

\section{Evaluation Diagnostics and Experimental Extensions}
\subsection{Failure Taxonomy}

Structured output allows for more precise failure analysis than a single free, text score of the output. errors should, at minimum, be assigned to six non, exclusive categories: \emph{serialization failure} occurs when the response is not parseable; \emph{schema failure} occurs when JSON is syntactically valid but keys, types, or required values are still missing; a A \emph{transition failure} occurs when the causal delta isn‘t a consequence of the action and prior state; a state, \emph{state-coherence failure} occurs when the effects don‘t align with the prior state (duplicates removed, certain persistent facts are lost, certain facts aren‘t supported, etc); a \emph{control failure} when reward or termination is inconsistent with the predicted state; finally, an \emph{evidence failure} occurs in grounded deployments when a valid transition isn‘t supported by retrieved sources.

This classification should be evaluated at record level as well as at domain level. One example can include a combination of another error class; for example, a valid JSON instance can have a proper reward field but an unsupported state update. Only time perfect exact match is reported we do not distinguish between different types of mechanisms. To analyze if the model is combining transition rules that can be applied to a wide variety of situations, or is simply learning the most common templates of its environment, domain stratified confusion matrices and representative set of errors cases are required.

Automated verifier can also introduce new risks. The most relevant example is that if the given evaluation is only assigned to precise fields, the model may learn policy formatting, find exploits due to lack of normalization, or maximize easily writable fields and not have to worry too much about writing challenging transition content which is more important. This is a misspecification of the reward structure, not mode behavior. 
\subsection{Required Ablation Matrix}
Fig.3  compares a base model with the final system after applying SFT and RL. That comparison shows an end-to-end gain especialy in JSON validiry, Schema validity, and done flag accuracy  metrics. For the JSON, Casual delta, and Resulting state indicators weren't available during the study, so authors put 0, however we believe that there's precise numbers certainly. 

Table~\ref{tab:ablations} outlines the minimum ablation matrix for the following experiment. We include the SFT, only condition, since it isolates the gains from supervised task learning with regard to the baseline. We include the constrained decoding condition, since it is capable of improving the JSON or schema validity independently from the transition semantics. We also add in a verifier, at, inference condition, since it is capable of providing the candidate generation and selection without the policy update.

\begin{table}[t]
\caption{Minimum Ablation Matrix for a Reproducible Rerun}
\label{tab:ablations}
\centering
\scriptsize
\begin{tabular}{@{}p{0.26\columnwidth}p{0.67\columnwidth}@{}}
\toprule
\textbf{Condition} & \textbf{Question answered} \\
\midrule
Base, prompt only & How much structure and transition ability exists before adaptation? \\
Base + constrained decoding & How much validity gain is purely syntactic? \\
QLoRA SFT & What is learned from curated demonstrations alone? \\
QLoRA SFT + verifier selection & Does inference-time verification improve exact state prediction? \\
QLoRA SFT + RLVR/GRPO & What is the incremental policy-learning gain over SFT? \\
SFT + RL + constrained decoding & What is the strongest defense-in-depth configuration? \\
Ablated reward fields & Which verifier components drive each reported metric? \\
\bottomrule
\end{tabular}
\end{table}

\subsection{Generalization and Stress Tests}
Random record splitting can overestimate generalization when templates recur across train and test. A stronger design should include three additional partitions. First, a \emph{held-out environment} split tests transfer to an unseen state vocabulary and rule set. Second, a \emph{held-out composition} split combines familiar rules in unfamiliar sequences. Third, a \emph{counterfactual stress} split minimally changes a precondition, entity, quantity, or temporal relation so that superficial pattern completion produces the wrong state.

Robustness testing should also vary serialization order, irrelevant context, paraphrase, contradictory instructions, missing preconditions, and incomplete evidence. For each perturbation, the system should either preserve the correct transition or abstain. In a regulated setting, a calibrated refusal is often preferable to a fluent unsupported answer. The stress suite should therefore report both correctness and coverage, making the trade-off between answered cases and residual error explicit.

\section{From Structured Reasoning to Legal Validation}
The current benchmark supports a deep-technology hypothesis: a language model can be adapted to emit structured, verifier-compatible state transitions. Therefore at this stage, it does not demonstrate reliable interpretation of court cases, case law, contractual language, or professional procedure. 

\subsection{Legal Task Taxonomy}
A first legal benchmark should include at least four task families:
\begin{enumerate}
    \item \textbf{Contract state tracking}: parties, clauses, obligations, exceptions, notice periods, renewals, and termination conditions;
    \item \textbf{Procedural timelines}: events, deadlines, jurisdiction-dependent rules, and completion status;
    \item \textbf{Compliance workflows}: applicable requirement, evidence, gap, remediation action, and reviewer approval; and
    \item \textbf{Evidence-grounded research}: proposition, source passage, jurisdiction, temporal validity, confidence, and abstention.
\end{enumerate}
LegalBench can inform reasoning-task coverage \cite{guha2023legalbench}, while CUAD provides a reference for expert-annotated contract review \cite{hendrycks2021cuad}. For large-scale deployment, task construction must use qualified legal professionals and versioned primary sources, since hallucination and errors in this kind of high-risk environments aren't acceptable.

\subsection{Evidence and Calibration Metrics}
In regulated workflows, output correctness is only one dimension. The validation suite should also measure citation precision, citation recall, support entailment, retrieval coverage, confidence calibration, selective risk under abstention, and consistency across repeated runs. Neural-network confidence scores are often miscalibrated, and temperature scaling provides a useful baseline for post-hoc calibration \cite{guo2017calibration}. Calibration should be assessed with expected calibration error, Brier score, and risk-coverage curves rather than a single confidence threshold.

\begin{table}[t]
\caption{Recommended Legal Validation Dimensions}
\label{tab:legalmetrics}
\centering
\scriptsize
\begin{tabular}{@{}p{0.25\columnwidth}p{0.68\columnwidth}@{}}
\toprule
\textbf{Dimension} & \textbf{Required evidence} \\
\midrule
State coherence & Entity, date, obligation, exception, and termination consistency \\
Grounding & Citation precision/recall, passage support, source freshness \\
Uncertainty & ECE, Brier score, risk-coverage, abstention quality \\
Robustness & Paraphrase, missing evidence, conflicting evidence, adversarial clauses \\
Reproducibility & Model hash, data manifest, seeds, prompts, verifier version \\
Operations & Latency, throughput, VRAM, failure recovery, audit-log completeness \\
Human oversight & Reviewer agreement, escalation rate, time saved, error severity \\
\bottomrule
\end{tabular}
\end{table}

\subsection{Release Gates}
\begin{figure}[t]
\centering
\begin{tikzpicture}[
node distance=4.2mm,
box/.style={draw=black, rounded corners=1pt, align=center, minimum height=7.5mm, text width=0.86\columnwidth, fill=grayC, font=\scriptsize},
line/.style={-Latex, thick}
]
\node[box] (g1) {Gate 1: versioned legal benchmark and data-rights register};
\node[box, below=of g1] (g2) {Gate 2: blind external legal review and error-severity analysis};
\node[box, below=of g2] (g3) {Gate 3: evidence grounding, calibration, and abstention thresholds};
\node[box, below=of g3] (g4) {Gate 4: GDPR, security, audit logging, and human-oversight controls};
\node[box, below=of g4, fill=grayB!45] (g5) {Gate 5: monitored paid pilot with rollback and incident process};
\draw[line] (g1) -- (g2);
\draw[line] (g2) -- (g3);
\draw[line] (g3) -- (g4);
\draw[line] (g4) -- (g5);
\end{tikzpicture}
\caption{Recommended release gates from internal benchmark to monitored legal pilot.}
\label{fig:gates}
\end{figure}
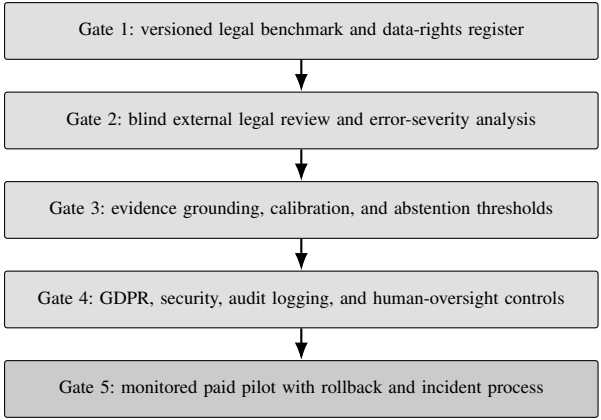

Fig.~\ref{fig:gates} defines a staged release process. No paid pilot should progress from assistive review to production use on the basis of aggregate accuracy alone. Each gate should have pre-registered pass/fail criteria, documented failure analysis, and independent expert sign-off.

\section{Sovereign Deployment and Governance}
The proposed system is intended for on-premise or customer-controlled private deployment. This design addresses confidentiality and procurement requirements by keeping documents, prompts, outputs, and audit logs within an approved boundary.

A production legal architecture should integrate retrieval from customer-controlled repositories, source pinning, model-version pinning, role-based access control, tamper-evident logs, retention policies, and explicit human approval. Retrieval can reduce reliance on parametric memory, but it could introduces its own risk modes, including missing documents, stale sources, incorrect ranking, and prompt injection. Thus, every generated legal proposition should be traceable to the evidence set used at inference time.

The governance design should be aligned with the General Data Protection Regulation and the EU Artificial Intelligence Act \cite{gdpr2016,aiact2024}. Relevant engineering practices include data minimization, purpose limitation, local processing by default, documented provider/deployer roles, human oversight, technical documentation, monitoring, and incident response. Model cards can provide a concise record of intended use, evaluation conditions, limitations, and excluded use cases \cite{mitchell2019modelcards}. SerenAI should be positioned as decision support for qualified professionals, not autonomous legal advice to consumers.

\section{Operational Threat Model and Assurance Evidence}
A sovereign deployment reduces dependence on third-party cloud services, but it does not remove model, retrieval, or application-layer risk. The relevant security boundary includes the model weights and adapters, customer documents, retrieval index, system prompts, tool permissions, generated outputs, and audit logs. Assurance should therefore be based on explicit threat scenarios and testable controls rather than the deployment location alone.

\subsection{Untrusted Retrieved Content}
A document-retrieval system necessarily processes content that may be erroneous, adversarial, or outside the authority of the user. Indirect prompt injection exploits the ambiguity between instructions and retrieved data, allowing malicious text embedded in a document to influence an LLM-integrated application \cite{greshake2023promptinjection}. For SerenAI, retrieved material should be treated as evidence, never as executable instruction. The runtime should preserve separate instruction and evidence channels, strip or quarantine active content, restrict tool calls through deterministic policies, and record which passages influenced each output. A red-team suite should include injected instructions in contracts, email chains, annexes, scanned text, and metadata.

\subsection{Confidentiality and Data Leakage}
On-premise inference limits routine data transfer, but confidentiality can still fail through access-control errors, logs, backups, generated text, or model memorization. Prior work has demonstrated that language models can expose memorized training examples under adversarial querying \cite{carlini2021extracting}. Customer documents should therefore be excluded from general model training by default. Where customer-specific tuning is contractually authorized, the resulting adapter should be isolated to that customer, accompanied by a data manifest, and tested for memorization and extraction risk. Logs should minimize raw personal data, use encryption and retention limits, and support access review.

\subsection{Evidence Freshness and Jurisdiction Drift}
Legal validity depends on jurisdiction, effective date, procedural posture, and source hierarchy. A citation can be authentic yet inapplicable because it is superseded, non-binding, or from the wrong jurisdiction. The retrieval layer should store source type, court or authority, publication date, effective interval, and version. The output schema for legal research should include jurisdiction and temporal-validity fields, while the verifier should reject conclusions that lack authoritative support. Periodic index refresh and source-revocation tests are required for operational maintenance.

\subsection{Automation Bias and Human Oversight}
Structured output can create an appearance of certainty even when the state prediction is wrong. Interfaces should avoid presenting a binary reward or termination field as a substitute for professional judgment. Reviewers should see the source passages, confidence or abstention signal, model version, and unresolved conflicts. High-severity actions, including deadline calculation, filing decisions, compliance approval, or contract termination, should require explicit human confirmation. Oversight effectiveness should be measured through reviewer agreement, detection of seeded errors, escalation behavior, and residual error severity rather than merely user satisfaction.

\begin{table}[t]
\caption{Operational Threats, Controls, and Audit Evidence}
\label{tab:threats}
\centering
\tiny
\setlength{\tabcolsep}{2pt}
\begin{tabular}{@{}p{0.20\columnwidth}p{0.36\columnwidth}p{0.34\columnwidth}@{}}
\toprule
\textbf{Threat} & \textbf{Primary controls} & \textbf{Required evidence} \\
\midrule
Indirect injection & Instruction/evidence separation, tool allow-list, content sanitization & Red-team corpus, tool-call logs, attack success rate \\
Cross-user leakage & Tenant isolation, RBAC, encryption, scoped indices & Penetration test, access review, isolation tests \\
Memory leakage & No default customer-data training, adapter isolation, output filtering & Extraction audit, training-data manifest \\
Stale authority & Versioned sources, jurisdiction metadata, revocation process & Freshness report, source lineage, temporal tests \\
Verifier bypass & Canonical parser, adversarial verifier tests, independent semantic checks & Unit-test coverage, failure cases, verifier hash \\
Automation bias & Mandatory review for high-severity actions, visible evidence and abstention & Reviewer study, seeded-error detection, escalation rate \\
\bottomrule
\end{tabular}
\end{table}

These points do not invalidate the reported experiment; they define the evidence required to make it scientifically reproducible. The most valuable next experiment is a fully logged rerun with a fixed model revision, public or escrowed test manifest, SFT-only and SFT+RL ablation, three or more seeds, domain-stratified results, and an expert-reviewed legal benchmark.

\section{Conclusion}
SerenAI reframes a subset of professional reasoning as structured state-transition prediction. Its four-field output contract enables automatic verification and provides a practical interface between language-model adaptation and regulated workflow controls. The reported internal results show substantial gains over an 8B base model in schema validity, transition exact match, reward prediction, and termination prediction. The evidence is sufficient to motivate further research, but not to claim it as ready legal reasoning tool. A defensible path forward combines reproducible training, legal-domain benchmarking, evidence-grounded retrieval, calibration, human oversight, and sovereign deployment. If these validation requirements are met, verifier-guided structured models may offer a useful reasoning layer for high-stakes workflows where auditability is as important as fluency.

\section*{Acknowledgment}
The author acknowledges the Collision Technologies team members for their contributions to the broader SerenAI research and product program. Collision Technologies keeps all rights to not disclose used data, training workflow and source codes.

\balance

\end{document}